\documentclass[conference]{IEEEtran}
\IEEEoverridecommandlockouts
\usepackage{cite}
\usepackage{amsmath,amssymb,amsfonts}
\usepackage{algorithmicx}
\usepackage{graphicx}
\usepackage{textcomp}
\usepackage{makecell}
\usepackage{algorithm}
\usepackage{algpseudocode}
\usepackage[dvipsnames]{xcolor}
\usepackage{textcomp}  
\usepackage{tabularx}

\newcommand{\tildecorrect}{\raisebox{0.5ex}{\texttildelow}}
\usepackage[hidelinks]{hyperref}
\hypersetup{
    colorlinks,
    linkcolor={red},
    citecolor={blue},
    urlcolor={blue}
}

\usepackage{booktabs}
\usepackage{caption}
\def\BibTeX{{\rm B\kern-.05em{\sc i\kern-.025em b}\kern-.08em
    T\kern-.1667em\lower.7ex\hbox{E}\kern-.125emX}}
\begin{document}

\title{Optimizing Methane Detection On Board Satellites: Speed, Accuracy, and Low-Power Solutions for Resource-Constrained Hardware\\
}
\author{\IEEEauthorblockN{1\textsuperscript{st} Jonáš Herec}
\IEEEauthorblockA{
\textit{Zaitra s.r.o.}\\
Brno, Czech Republic \\
jonas.herec@zaitra.io}
\and
\IEEEauthorblockN{2\textsuperscript{nd} Vít Růžička}
\IEEEauthorblockA{\textit{University of Oxford} \\
Oxford, United Kingdom \\
previtus@gmail.com}
\and
\IEEEauthorblockN{3\textsuperscript{rd} Rado Pitoňák}
\IEEEauthorblockA{
\textit{Zaitra s.r.o.}\\
Brno, Czech Republic \\}
}

\maketitle

\begin{abstract}
Methane is a potent greenhouse gas, and detecting its leaks early via hyperspectral satellite imagery can help mitigate climate change. Meanwhile, many existing missions operate in manual tasking regimes only, thus missing potential events of interest. To overcome slow downlink rates cost-effectively, onboard detection is a viable solution. However, traditional methane enhancement methods are too computationally demanding for resource-limited onboard hardware. This work accelerates methane detection by focusing on efficient, low-power algorithms. We test fast target detection methods (ACE, CEM) that have not been previously used for methane detection and propose a Mag1c-SAS -- a significantly faster variant of the current state-of-the-art algorithm for methane detection: Mag1c. To explore their true detection potential, we integrate them with a machine learning model (U-Net, LinkNet). Our results identify two promising candidates (Mag1c-SAS and CEM), both acceptably accurate for the detection of strong plumes and computationally efficient enough for onboard deployment: one optimized more for accuracy, the other more for speed, achieving up to \tildecorrect100× and \tildecorrect230× faster computation than original Mag1c on resource-limited hardware. Additionally, we propose and evaluate three band selection strategies. One of them can outperform the method traditionally used in the field while using fewer channels, leading to even faster processing without compromising accuracy. This research lays the foundation for future advancements in onboard methane detection with minimal hardware requirements, improving timely data delivery. The produced code, data, and models are open-sourced and can be accessed from \url{https://github.com/zaitra/methane-filters-benchmark}.
\end{abstract}


\section{Introduction}

Methane is the second most significant greenhouse gas contributing to global warming after $\text{CO}_2$~\cite{kuylenstierna2021global_UN}. While $\text{CO}_2$ is more prevalent, methane is more potent, mainly in the short term (84-87 times more than $\text{CO}_2$ over a 20-year period~\cite{Intergovernmental}). A significant share of methane emissions in the oil and gas industry originates from episodic ultra-emission events, often caused by equipment failures and other unpredictable factors~\cite{lauvaux2022global}.

These events could be detected and mitigated early with the use of satellite monitoring, but it is a challenging task. While methane has a distinct spectral signature with its strongest absorption features in the \tildecorrect2100–2500 nm range, the change in measured radiance is too subtle and cannot be detected by the human eye. However, it can be seen in data from current imaging spectroscopy sensors such as NASA's AVIRIS-NG or EMIT~\cite{green2022_EMIT} and from the sensors of planned missions such as NASA's SBG~\cite{cawse2021nasa_SBG}, or the ESA's CHIME~\cite{rast2021copernicus_CHIME}. These missions carry hyperspectral instruments that nearly continuously capture light across hundreds of spectral channels, allowing to leverage methane’s spectral signature for its detection.


The high number of channels is advantageous, but it also hinders the timely delivery of data. Downlinking from satellites is often slow, as data is transmitted via limited-bandwidth radio communication only a few times per day when the satellite passes over a ground station. Increasing transmission speed would significantly raise costs, not only due to the added weight and expense of more powerful satellite antennas, but also because it would require a larger network of ground stations.
The approach of many currently flown satellites (for example, PRISMA or EnMAP) is to observe only over manually selected regions, which severely reduces the number of potentially captured methane leak events.

To address this, algorithms and machine learning (ML) models can be employed on board, as was explored for disaster events in \cite{RaVAEn}, and specifically for methane detection in \cite{Ruzicka2025HyperspectralViTs}. However, since the goal is to reduce the cost and improve the timely delivery of salient data, these models must be designed to fit the constraints of the mission without compromising its objectives. Given the high power requirements of GPUs and their ongoing adaptation for the space environment, the key could be developing efficient algorithms that can run on commonly used solutions~\cite{nasasmallsats} -- lower-power CPUs, or FPGAs~\cite{guerrieri2018fpgas_for_space}.

The key part of the pipeline is the computation of the methane enhancement product, which generates a 2D image representing the detection of the methane spectral signature from the 3D hyperspectral cube. This process effectively reduces the number of channels from dozens or hundreds to just one, which can then be used for methane detection by thresholding or by an efficient machine learning model, enabling real-time methane detection.

\begin{figure}[ht!]
\centerline{\includegraphics[width=0.5\textwidth]{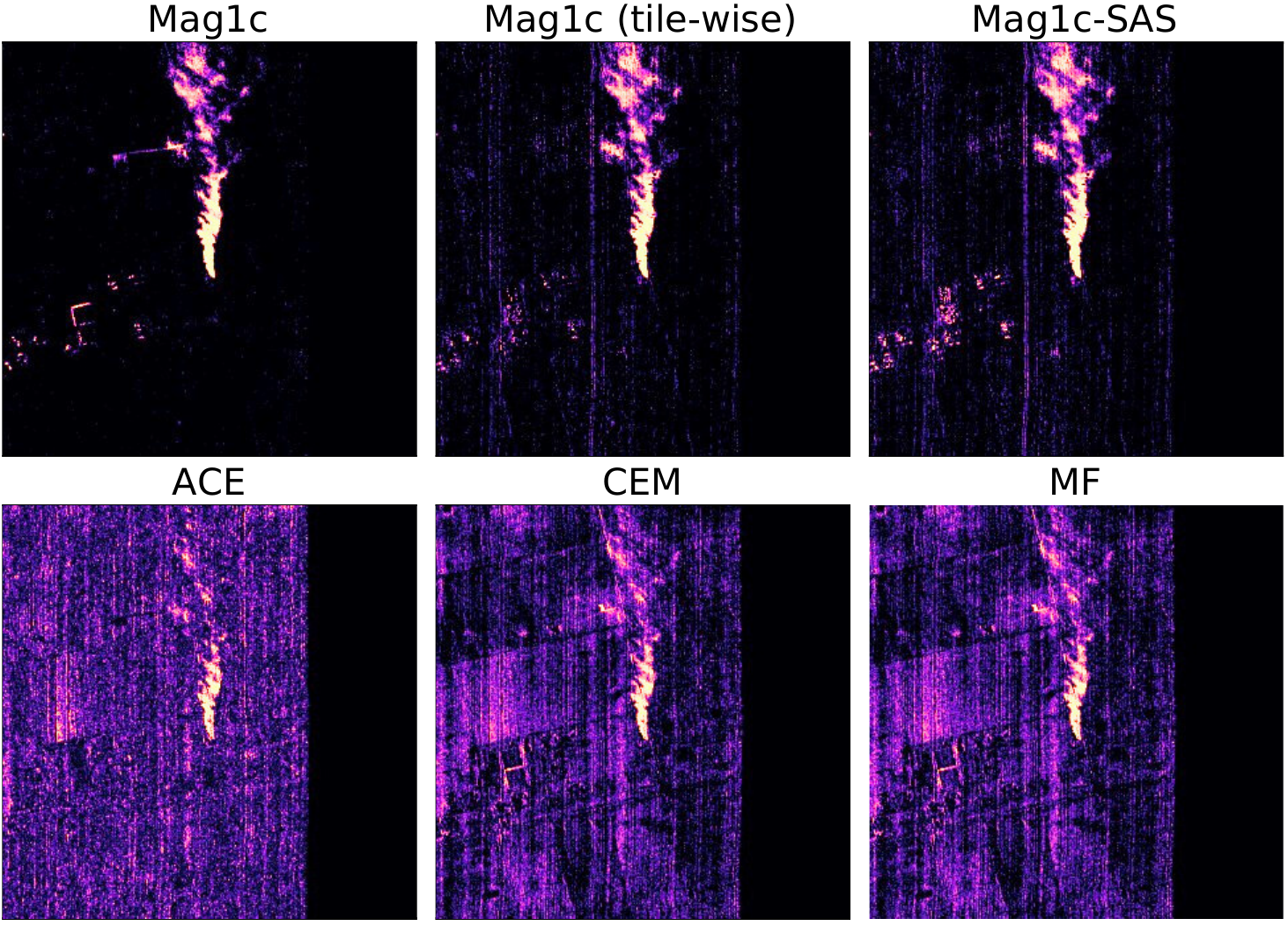}}
\caption{Visualisation of a strong methane plume event showing all methane enhancement products used in this study. Note that visual scaling was adjusted manually to aid legibility.}
\label{direct}
\vspace{-6mm}
\end{figure}

Typically used product is the Matched filter~\cite{tiemann2024machine}, with Matched filter with Albedo correction and reweiGhted L1 sparsity Code (Mag1c)~\cite{mag1c} being a state-of-the-art variant. These products are calculated for each column of the input image; in Mag1c, this is further repeated in several iterations. We argue that these products are prohibitively slow for onboard deployment. Therefore, we explore faster and simpler variants of these methane enhancement products, where, besides other optimizations, we compute the whole filter at once instead of column-wise, to further speed up the process.

Since the number of bands is a crucial parameter for these algorithms, we experiment with newly proposed band selection strategies to determine if the runtime can be further reduced by selecting fewer bands without compromising accuracy. We trade in the quality of the products for speed, but this can be mitigated by lightweight machine learning models, which refine the methane product, similarly to \cite{jongaramrungruang_methanet_2022, ruuvzivcka2023semantic}. Also, we focus on more easily detectable large plumes, which are particularly relevant, as their impact is disproportionately larger than that of smaller leaks \cite{duren_californias_2019} and is thus more critical for onboard detection.

Namely, our contributions are: 
\begin{itemize}
    \item We propose a new methane product called ``Mag1c Sped up with Additional Sparsity'' (\textbf{Mag1c-SAS}), which is almost 100x faster than the original Mag1c applied on columns, with an acceptable drop in accuracy.
    \item We benchmark three newly proposed band selection strategies and test the influence of the number of channels on accuracy and speed.
    \item We investigate various target detection methods, including the Constrained Energy Minimization (CEM) method \cite{harsanyi1993detection}, which is nearly 230× faster than the original Mag1c applied to columns.
    \item Finally, we explore the capabilities of the ML model to refine these products, where we consistently observe improvements in the F1 score by up to \tildecorrect20 points for strong plumes.
\end{itemize}
\section{Methodology}

\subsection{Methane enhancement products}
\label{products}

Figure \ref{direct} shows a selection of methane enhancement products explored in this work. 
A widely used method for methane detection is the matched filter~\cite{manolakis2003hyperspectral}, often implemented with various modifications. Its primary goal is to identify and amplify the methane signal while suppressing background noise. This is accomplished by comparing the spectral signature to deviations from the band-wise means at each hyperspectral pixel while considering the background characteristics modeled by the band’s variance-covariance matrix.

Here, we use the \textbf{matched filter} as defined for classical target detection tasks in \cite{matchedfilteroriginal}:
\[
y_i = \frac{(x_i - \mu)^\top \, C^{-1} (t - \mu)}{(t - \mu)^\top C^{-1} (t - \mu)}
\]
\noindent where:
\begin{itemize}
  \item \( x_i \in \mathbb{R}^{p} \) is the hyperspectral pixel with dimensionality \( p \) (number of bands).
  \item \( t \in \mathbb{R}^{p} \) is the target spectrum to detect.
  \item \( \mu \in \mathbb{R}^{p} \) is vector of band-wise means, i.e.,
\[
\mu = \frac{1}{N} \sum_{i=1}^{N} x_i,
\]
  \item \( C \in \mathbb{R}^{p \times p} \) is the covariance matrix of the mean-centered data:
\[
C = \frac{1}{N - 1} \sum_{i=1}^{N} (x_i - \mu)^\top(x_i - \mu)
\]
  \item \(y_i\) is the matched filter response at pixel: a scalar, indicating how well this pixel matches the target spectrum.
\end{itemize}

Very similar to the matched filter is the \textbf{Constrained Energy Minimization (CEM)} detector~\cite{harsanyi1993detection}. While it originates from a different theoretical background than MF, it arrives at an almost identical filter definition, but without the whitening (mean subtraction) of the data. The CEM detector is defined as:
\[
y_i = \frac{x_i^\top \, K^{-1} t}{t^\top K^{-1} t}
\]
\noindent where:
\begin{itemize}
\item $K$ (often referred to as sample correlation matrix or auto-correlation matrix) is similar to $C$, but it is computed directly from the raw (non-centered) data:
\[
K = \frac{1}{N - 1} \sum_{i=1}^{N} x_i^\top x_i
\]
\end{itemize}
The last classic target detection method used in this study is the \textbf{Adaptive Cosine Estimator (ACE)}~\cite{10.1117/12.818790}. Defined as follows:
\[
y_i = \frac{\left[(x_i - \mu)^\top C^{-1} (t - \mu)\right]^2}{\left[(t - \mu)^\top C^{-1} (t - \mu)\right] \cdot \left[(x_i - \mu)^\top C^{-1} (x_i - \mu)\right]}
\]

Again, its core principles are very similar to MF and CEM, but it has additional normalization by the value of the pixel itself. Thus, two pixels (vectors) with the same spectral response (direction) but different brightness (magnitude) will receive the same score.
 
While these methods are typically used as baselines in target detection research, they could be ideal candidates for this application. This is because more advanced target detection methods often offer only a slight increase in accuracy but result in a significant increase in processing time~\cite{slightaccuracyformorespeed}.

For remote sensing imagery, the matched filter was particularly adapted. In these applications~\cite{mf1,mf2}, the algorithm is applied on each column separately, so the key parameters (\(\mu, C\)) are estimated separately for each sensor in a push-broom instrument. Due to calibration and manufacturing differences of the sensors, this approach ensures more robust and accurate estimation of the background characteristics, improving the accuracy and eliminating artifacts like striping in the final output.

However, this involves the calculation of band means, the construction and inversion of the variance-covariance matrix for each column, which can be highly computationally demanding. Additionally, one of the most widely used and accurate methods, Mag1c~\cite{mag1c}, further increases computational complexity by performing an iterative calculation of the filter. There, the means and variance-covariance matrix are iteratively re-estimated from data after removing the methane enhancement from the previous iteration, as estimating them from methane-contaminated data inhibits the filter's detection capabilities. 

The original Mag1c pseudocode is shown in Algorithm~\ref{alg:spatialrwl1mf}. Its input arguments are: the hyperspectral datacube $D \in \mathbb{R}^{H \times W \times p}$, where $H$ is an image height and $W$ an image width and $p$ the number of bands; a methane spectrum $s \in \mathbb{R}^{p}$; and $N_{\mathrm{iter}}$, an integer indicating the number of iterations. Detection, however, is performed independently for each column $L \in \mathbb{R}^{H \times p}$. On row \ref{core-principle} is its core, which when expanded is:
\[
\alpha_i^{k} = \max \left( 
\frac{
\left( \boldsymbol{L}_i - \boldsymbol{\mu}^k \right)^T 
{\boldsymbol{C}^k}^{-1} 
\left( \boldsymbol{\mu}^{k} \odot \boldsymbol{s} \right) 
- w_i^k
}{
r_i 
\left( 
\left( \boldsymbol{\mu}^{k} \odot \boldsymbol{s} \right)^T 
{\boldsymbol{C}^k}^{-1} 
\left( \boldsymbol{\mu}^{k} \odot \boldsymbol{s} \right) 
\right) 
},\ 0 
\right)
\]

If we set the sparsity-enforcing term $w_i^k = 0$ and the albedo correction term $r_i = 1$, the resulting formulation closely resembles the previously described matched filter. However, it is passed through a ReLU function, and instead of subtracting the mean values from the target spectrum, they are multiplied by it. This models the methane influence on the observed values as multiplicative rather than additive, as seen in the previously defined target detection methods, and it also allows the output to be expressed in units of ppm/m.

However, even though the core principle is similar, the algorithm results differ mainly because of 3 changes:
\begin{itemize}
\item \textbf{1.} It uses an albedo correction term $r_i = \frac{
                                        \boldsymbol{L}_i^T \boldsymbol{\mu}
                                    }{
                                        \boldsymbol{\mu}^T \boldsymbol{\mu}
                                    }$, which is used in the product denominator. Methane detections, which are stronger due to the greater reflectability of their background (brighter pixels), are thus normalized by a greater value.
\item \textbf{2.} It uses a sparsity-enforcing term $\boldsymbol{w}^k = \frac{1}{\textcolor{red}{r} (\boldsymbol{\alpha}^{k-1} + \epsilon)}$, which is subtracted from the methane detection in the numerator. Combined with the ReLU function and the iterative nature of Mag1c, this term pushes small and likely false-positive detections to zero.
\item \textbf{3.} As the background characteristics (\(\mu, C\)) should be methane free to enable correct methane signal amplification, the filter is computed in $N_{\mathrm{iter}}$ iterations, where in each iteration, the previous methane detection is subtracted from the pixel during the parameters computation, as seen in rows \ref{first-row}-\ref{last-row}.
\end{itemize}
Besides the original, we explore two accelerated variants of the Mag1c product:

\begin{itemize}
\item \textbf{Mag1c (tile-wise):} Instead of applying Mag1c column-wise, we apply it to the whole tile at once to speed up the computation on the low-power CPU.
\item \textbf{Mag1c-SAS:} The computation is also performed over the entire tile but split into two stages. The first stage involves computationally heavy iterative parameter estimation, but this is run on only a small fraction ($f = 1\,\%$) of the tile. In the second stage, the Mag1c-SAS product is computed across the full tile using the parameters estimated in the first stage. Although iterative computation is still necessary in the second stage to enforce sparsity, the matched filter is computed only once here; as in each iteration, only the sparsity-enforcing term ($\boldsymbol{w}^k$) is recalculated, normalized, and subtracted from the filter output. These changes yield a slightly less precise but substantially faster algorithm, making it well-suited for low-power CPUs. The full customized algorithm is detailed in Algorithm~\ref{alg:mag1c-sas}. \textit{The $f$-subset is selected using uniformly spaced steps over the flattened tile to ensure both representativeness and reproducibility.}
\end{itemize}

\begin{figure}[H]
\centerline{\includegraphics[width=0.5\textwidth]{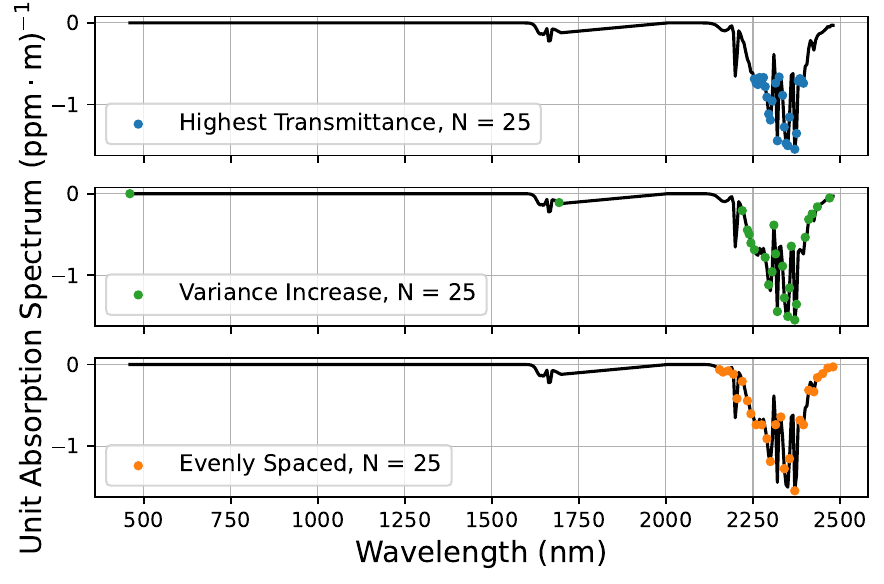}}
\caption{Comparison of selection strategies for the STARCOP data target spectrum.}
\label{selection-strategy}
\vspace{-6mm}
\end{figure}

\begin{figure}[ht!]
    \vspace{-5mm}
    \centering
    \begin{algorithm}[H]
        \caption{Description of the original Mag1c. \textcolor{red}{Red} indicates content added or modified in this article. Each change was done for clarity or to reflect important details present in the code but omitted from the original paper (see Figure 1 in \cite{mag1c}).}
        \label{alg:spatialrwl1mf}
        \begin{algorithmic}[1]
        
            {\color{red}
            \Procedure{Mag1c}{$\boldsymbol{D}$, $\boldsymbol{s}$, $N_{\mathrm{iter}}$}
                \State Initialize $\boldsymbol{A}$ as $H \times W$ matrix
                \ForAll{$j$} 
                    \State Extract column $j$ from $\boldsymbol{D}$: $\boldsymbol{d}_j \in \mathbb{R}^{H \times p}$
                    \State $\boldsymbol{\alpha}_j \gets \textsc{AlbedoReWeightL1Filter}(\boldsymbol{d}_j, \boldsymbol{s}, N_{\mathrm{iter}})$
                    \State $\boldsymbol{A}[:, j] \gets \boldsymbol{\alpha}_j$
                \EndFor
                \State \textbf{return} $\boldsymbol{A}$
            \EndProcedure}
            \Procedure{AlbedoReWeightL1Filter}{\textcolor{red}{$\boldsymbol{L} $}, $\boldsymbol{s}$, $N_{\mathrm{iter}}$}
            \State Initialize $\boldsymbol{\mu}^0 = \frac{1}{N} \sum_{i}^{N} \boldsymbol{L}_i$
            \State Initialize $\boldsymbol{C}^0 = 
                                \frac{1}{N} \sum_{i}^{N} 
                                \left(
                                    \boldsymbol{L}_i - \boldsymbol{\mu}^0 
                                \right)^T 
                                \left(
                                    \boldsymbol{L}_i   - \boldsymbol{\mu}^0 
                                \right)$
            \ForAll{$i$}
                \State Set $r_i = \frac{
                                        \boldsymbol{L}_i^T \boldsymbol{\mu}
                                    }{
                                        \boldsymbol{\mu}^T \boldsymbol{\mu}
                                    }$
                \State Initialize $ \alpha^0_i = 
                                        \frac{
                                            \left(
                                                \boldsymbol{L}_i - \boldsymbol{\mu}^0
                                            \right)^T 
                                            {\boldsymbol{C}^0}^{-1} 
                                            \left(
                                                \boldsymbol{\mu}^0 \odot \boldsymbol{s} 
                                            \right)
                                        }{
                                            r_i
                                            \left(
                                                \boldsymbol{\mu}^0 \odot \boldsymbol{s}
                                            \right)^T 
                                            {\boldsymbol{C}^0}^{-1} 
                                            \left(
                                                \boldsymbol{\mu}^0 \odot \boldsymbol{s}
                                            \right)
                                        } $ 
            \EndFor
            \For{$k = 1$ \textbf{to} $N_{\mathrm{iter}}$} \color{red}
                \State \color{black} $\boldsymbol{w}^k = \frac{1}{\textcolor{red}{r} (\boldsymbol{\alpha}^{k-1} + \epsilon)}$
                \State $\boldsymbol{\mu}^k = 
                            \frac{1}{N} \sum_{i}^{N} 
                                \left(
                                    \boldsymbol{L}_i - r_i \alpha^{k-1}_i \boldsymbol{\mu}^{k-1} \odot \boldsymbol{s}   
                                \right)$ \label{first-row}
                \ForAll{$i$}
                    \State Let $\boldsymbol{d}_{Ci} = 
                        \boldsymbol{L}_i 
                        - r_i \alpha^{k-1}_i \boldsymbol{\mu}^{k} \odot \boldsymbol{s}  
                        - \boldsymbol{\mu}^k$
                \EndFor
                \State $\boldsymbol{C}^k = 
                            \frac{1}{N} \sum_{i}^{N} 
                                \boldsymbol{d}_{Ci}\boldsymbol{d}_{Ci}^T$ \label{last-row}
                \ForAll{$i$}
                    \color{red}\State $m = 
                                            \left(
                                                \boldsymbol{\mu}^{k} \odot \boldsymbol{s}
                                            \right)^T 
                                            {\boldsymbol{C}^k}^{-1} 
                                            \left(
                                                \boldsymbol{\mu}^{k} \odot \boldsymbol{s}
                                            \right)$
                    \State $m = m\text{.clamp(min=1)}$
                    \State \color{black}$\alpha_i^{k} = 
                                \max 
                                    \left(
                                        \frac{
                                            \left(
                                                \boldsymbol{L}_i - \boldsymbol{\mu}^k
                                            \right)^T 
                                            {\boldsymbol{C}^k}^{-1} 
                                            \left(
                                                \boldsymbol{\mu}^{k} \odot \boldsymbol{s}
                                            \right) 
                                            - w_i^k
                                        }{
                                            r_i 
                                            \textcolor{red}{m}
                                        }
                                    , 0
                                    \right)$ \label{core-principle}
                \EndFor
            \EndFor
            \State \textbf{return} $\boldsymbol{\alpha}^{N_{\mathrm{iter}}}$
            \EndProcedure
        \end{algorithmic}
    \end{algorithm}
\vspace{-8mm}
\end{figure}

\begin{figure}[ht!]
    \vspace{-5mm}
    \centering
    \begin{algorithm}[H]
        \caption{Mag1c-SAS algorithm. The computationally intensive parameter estimation is performed only on a small fraction of the data and then reused in a lightweight filter. Changes to the original Mag1c are highlighted as \textcolor{ForestGreen}{green}.}
        \label{alg:mag1c-sas}
        \begin{algorithmic}[1]
            \Procedure{Mag1c\textcolor{ForestGreen}{-SAS}}{$\boldsymbol{D}$, $\boldsymbol{s}$, $N_{\mathrm{iter}}$, \textcolor{ForestGreen}{$f=0.01$}}
                \color{ForestGreen}\State Reshape $\boldsymbol{D}$ into $H \cdot W \times p$ matrix
                \State Sample a fraction $f$ of $\boldsymbol{D}$ $\to$ $\boldsymbol{D}_{\text{sample}}$
                 \State $\boldsymbol{\mu}, \boldsymbol{t}, m \gets \textsc{ComputeParameters}(\boldsymbol{D}_{\text{sample}}, \boldsymbol{s}, N_{\mathrm{iter}})$
                \State $\boldsymbol{A} \gets \textsc{LightWeightFilter}(\boldsymbol{D}, N_{\mathrm{iter}}, \boldsymbol{\mu}, \boldsymbol{t}, m)$
                \State Reshape $\boldsymbol{A}$ into $H \times  W$ matrix
                \color{Black}
                \State \textbf{return} $\boldsymbol{A}$
            \EndProcedure
            \State
            \Procedure{\textcolor{ForestGreen}{ComputeParameters}}{$\boldsymbol{L} $, $\boldsymbol{s}$, $N_{\mathrm{iter}}$}
            \State Initialize $\boldsymbol{\mu}^0 = \frac{1}{N} \sum_{i}^{N} \boldsymbol{L}_i$
            \State Initialize $\boldsymbol{C}^0 = 
                                \frac{1}{N} \sum_{i}^{N} 
                                \left(
                                    \boldsymbol{L}_i - \boldsymbol{\mu}^0 
                                \right)^T 
                                \left(
                                    \boldsymbol{L}_i   - \boldsymbol{\mu}^0 
                                \right)$
            \ForAll{$i$}
                \State Set $r_i = \frac{
                                        \boldsymbol{L}_i^T \boldsymbol{\mu}
                                    }{
                                        \boldsymbol{\mu}^T \boldsymbol{\mu}
                                    }$
                \State Initialize $ \alpha^0_i = 
                                        \frac{
                                            \left(
                                                \boldsymbol{L}_i - \boldsymbol{\mu}^0
                                            \right)^T 
                                            {\boldsymbol{C}^0}^{-1} 
                                            \left(
                                                \boldsymbol{\mu}^0 \odot \boldsymbol{s} 
                                            \right)
                                        }{
                                            r_i
                                            \left(
                                                \boldsymbol{\mu}^0 \odot \boldsymbol{s}
                                            \right)^T 
                                            {\boldsymbol{C}^0}^{-1} 
                                            \left(
                                                \boldsymbol{\mu}^0 \odot \boldsymbol{s}
                                            \right)
                                        } $ 
            \EndFor
            \For{$k = 1$ \textbf{to} $N_{\mathrm{iter}}$} 
                \State $\boldsymbol{w}^k = \frac{1}{r (\boldsymbol{\alpha}^{k-1} + \epsilon)}$
                \State $\boldsymbol{\mu}^k = 
                            \frac{1}{N} \sum_{i}^{N} 
                                \left(
                                    \boldsymbol{L}_i - r_i \alpha^{k-1}_i \boldsymbol{\mu}^{k-1} \odot \boldsymbol{s}   
                                \right)$
                \ForAll{$i$}
                    \State Let $\boldsymbol{d}_{Ci} = 
                        \boldsymbol{L}_i 
                        - r_i \alpha^{k-1}_i \boldsymbol{\mu}^{k} \odot \boldsymbol{s}  
                        - \boldsymbol{\mu}^k$
                \EndFor
                \State $\boldsymbol{C}^k = 
                            \frac{1}{N} \sum_{i}^{N} 
                                \boldsymbol{d}_{Ci}\boldsymbol{d}_{Ci}^T$
                \ForAll{$i$}
                    \State $m = 
                                            \left(
                                                \boldsymbol{\mu}^{k} \odot \boldsymbol{s}
                                            \right)^T 
                                            {\boldsymbol{C}^k}^{-1} 
                                            \left(
                                                \boldsymbol{\mu}^{k} \odot \boldsymbol{s}
                                            \right)$
                    \State $m = m\text{.clamp(min=1)}$
                    \State $\alpha_i^{k} = 
                                \max 
                                    \left(
                                        \frac{
                                            \left(
                                                \boldsymbol{L}_i - \boldsymbol{\mu}^k
                                            \right)^T 
                                            {\boldsymbol{C}^k}^{-1} 
                                            \left(
                                                \boldsymbol{\mu}^{k} \odot \boldsymbol{s}
                                            \right) 
                                            - w_i^k
                                        }{
                                            r_i 
                                            m
                                        }
                                    , 0
                                    \right)$
                \EndFor
            \EndFor
            \color{ForestGreen}
            \State $\boldsymbol{t} = {\boldsymbol{C}^{N_{\mathrm{iter}}}}^{-1} 
    \left(
        \boldsymbol{\mu}^{N_{\mathrm{iter}}} \odot \boldsymbol{s}
    \right)$
            \State $m = \left(
                    \boldsymbol{\mu}^{N_{\mathrm{iter}}} \odot \boldsymbol{s}
                \right)^T 
                {\boldsymbol{t}}$
            \State \textbf{return} $\boldsymbol{\mu}^{N_{\mathrm{iter}}}$, $\boldsymbol{t}$, $m$
            \color{black}
            \EndProcedure
            
            \color{ForestGreen}\Procedure{LightweightFilter}{$\boldsymbol{L} $, $N_{\mathrm{iter}}$, $\boldsymbol{\mu}$, $\boldsymbol{t}$, $m$}
            \ForAll{$i$}
                \State $r_i = \frac{
                                        \boldsymbol{L}_i^T \boldsymbol{\mu}
                                    }{
                                        \boldsymbol{\mu}^T \boldsymbol{\mu}
                                    }$ \label{alg:albedostep}
                \State $m = m\text{.clamp(min=1)}$
                \State $ \alpha^0_i = 
                                        \frac{
                                            \left(
                                                \boldsymbol{L}_i - \boldsymbol{\mu}
                                            \right)^T 
                                            {\boldsymbol{t}}}{
                                            r_i m
                                        } $ 
            \EndFor
            \For{$k = 1$ \textbf{to} $N_{\mathrm{iter}}$} 
                \State $\boldsymbol{w}^k = \frac{1}{r ( \boldsymbol{\alpha}^{k-1} + \epsilon)}$
                \State $\alpha_i^{k} = 
                            \max 
                                \left(\alpha_i^{0}-
                                    \frac{w_i^k
                                    }{
                                        r_i m 
                                    }
                                , 0
                                \right)$
            \EndFor
            \State \textbf{return} $\boldsymbol{\alpha}^{N_{\mathrm{iter}}}$
            \EndProcedure
        \end{algorithmic}
    \end{algorithm}
\vspace{-10mm}
\end{figure}
\subsection{Data}
We use the STARCOP dataset created in \cite{ruuvzivcka2023semantic}. This dataset consists of 512x512 tiles of hyperspectral remote sensing images taken by the aerially flown sensor AVIRIS-NG~\cite{thorpe2016mapping_AVIRISNG} and their semantic segmentation labels of methane plumes. We use the so-called ``allbands'' variant of the dataset released by \cite{Ruzicka2025HyperspectralViTs}, which has 125 spectral bands (RGB + 122 channels in the methane-relevant range of 1573-2480nm). For training, the dataset of 3425 tiles is used; for evaluation, we rely on the STARCOP test set of 342 tiles (these are further stratified into weak and strong plume events using the size of the label).
\vspace{-1mm}

\subsection{Band selection}
A crucial hyperparameter is the number of bands used and the method used for their sampling. Generally, using more bands can result in better accuracy but also longer processing time. Therefore, finding the optimal strategy and number of bands is essential. We tested 3 strategies, visualized in Figure \ref{selection-strategy}, which involve selecting N bands based on:
\begin{itemize}
\item \textbf{Highest (absolute) transmittance:} Select the bands with the highest absolute transmittance, as these bands have the strongest methane signal.
\item \textbf{Variance increase:} Start with the band that has the highest transmittance, and select each subsequent band to maximize the variance in CH$_4$ transmittance. This allows for a more accurate approximation of the methane transmittance function.
\item \textbf{Evenly spaced in the main methane transmittance range:} Select bands between \tildecorrect2100-2500 nm with even spacing between them.
\end{itemize}


\subsection{Inference}\label{inference}
\label{post-processing}
For each testing run, the test set is transformed from the spectral bands into the methane enhancement product using the selected N bands. Then, two inference strategies are used:

\begin{itemize}
\item \textbf{Morphological baseline:} The product is first thresholded (each product uses a different experimentally found threshold value). Then, it is processed morphologically, starting with erosion followed by dilation, which reduces the salt and pepper noise.
\item \textbf{ML model:}
Inspired by \cite{ruuvzivcka2023semantic}, the inputs consist of RGB bands along with the methane enhancement product. The idea is that these products may contain noise or false positives (e.g., solar panels), which can be identified in the RGB bands. The training process is practically the same as in \cite{ruuvzivcka2023semantic}, except that different methane products are used. Namely, the training dataset is tiled into 128x128px patches with an overlap of 64px.
These tiles are sampled with the PyTorch WeightedRandomSampler to balance the number of tiles with and without methane leak events. For data augmentation, we use random rotations, horizontal and vertical flips. Finally, during training, we use the binary cross-entropy loss multiplied by the Mag1c product provided by \cite{ruuvzivcka2023semantic}. We highlight that this is used only during training to guide the model towards regions with strong methane plume events and also confounder areas with strong false signals in the computed methane products. All models were implemented via the Segmentation Models Pytorch library~\cite{SMP}, which allows for easy combination of SoTa encoders and decoders. Namely, we use:
    \begin{itemize}
    \item \textbf{U-Net:}
    As a primary model, we used a classic U-Net architecture \cite{ronneberger2015u} with the fast MobileNet-v2 encoder \cite{sandler2018mobilenetv2}, resulting in 6.6 million parameters and a 25.27\,MB size.
    \item \textbf{LinkNet:}
    As the secondary model, we use LinkNet \cite{chaurasia2017linknet}. It has an Encoder-Decoder-like architecture with skip connections similar to U-Net, but it is designed to be smaller and faster, mainly by using element-wise addition in skip connections instead of concatenation. We also used a very small version of MobileNetv3\cite{mobilenet3} encoder -- \texttt{timm-mobilenetv3\_small\_minimal\_100}, designed in the PyTorch Image Models (timm) library~\cite{timm}. This encoder-decoder combination resulted in 0.851 million parameters and 3.34\,MB size.
    \end{itemize}
\end{itemize}

\subsection{Computational environment}\label{computational-environment}
The models are trained on a high-performance computing (HPC) cluster with NVIDIA Tesla V100 GPUs. Typical training takes around 8 hours (depending on small IO fluctuations). 

The runtime is measured on the Raspberry Pi 3 B+ device with 1 GB of RAM and a 4-core Cortex-A53 @1.2-1.4\,GHz CPU. While the Raspberry Pi has been considered for use in space only for mission non-critical tasks by NASA~\cite{guertin2022raspberry}, it can be considered as an example of a low-compute device representative of computational power available on real satellite missions like Phi-Sat-1 \cite{giuffrida2021phi_sat_1_mission} or D-Orbit's ION-SCV satellites \cite{TrainOnBoard}. Also, the Cortex-A53 CPU is used, for example, in the space-qualified Xiphos Q8 board with a flight heritage of 100+ units~\cite{xiphos_q8}.

Although we applied various adjustments to decrease runtime, we kept the products in their native implementation format, so some intra-products speed differences can occur due to the library implementation details, as all Mag1c products, along with morphological baseline, are implemented in PyTorch, whereas the other products (ACE, CEM, MF) and their accelerated versions are implemented using NumPy and SciPy. The ML models were compiled using ONNX for a fixed input size of 4x512x512, and deployed using ONNX Runtime~\cite{onnxruntime}.

\section{Results - Methane segmentation}
\begin{figure*}[p!]
    \centering
    \includegraphics[width=\textwidth]{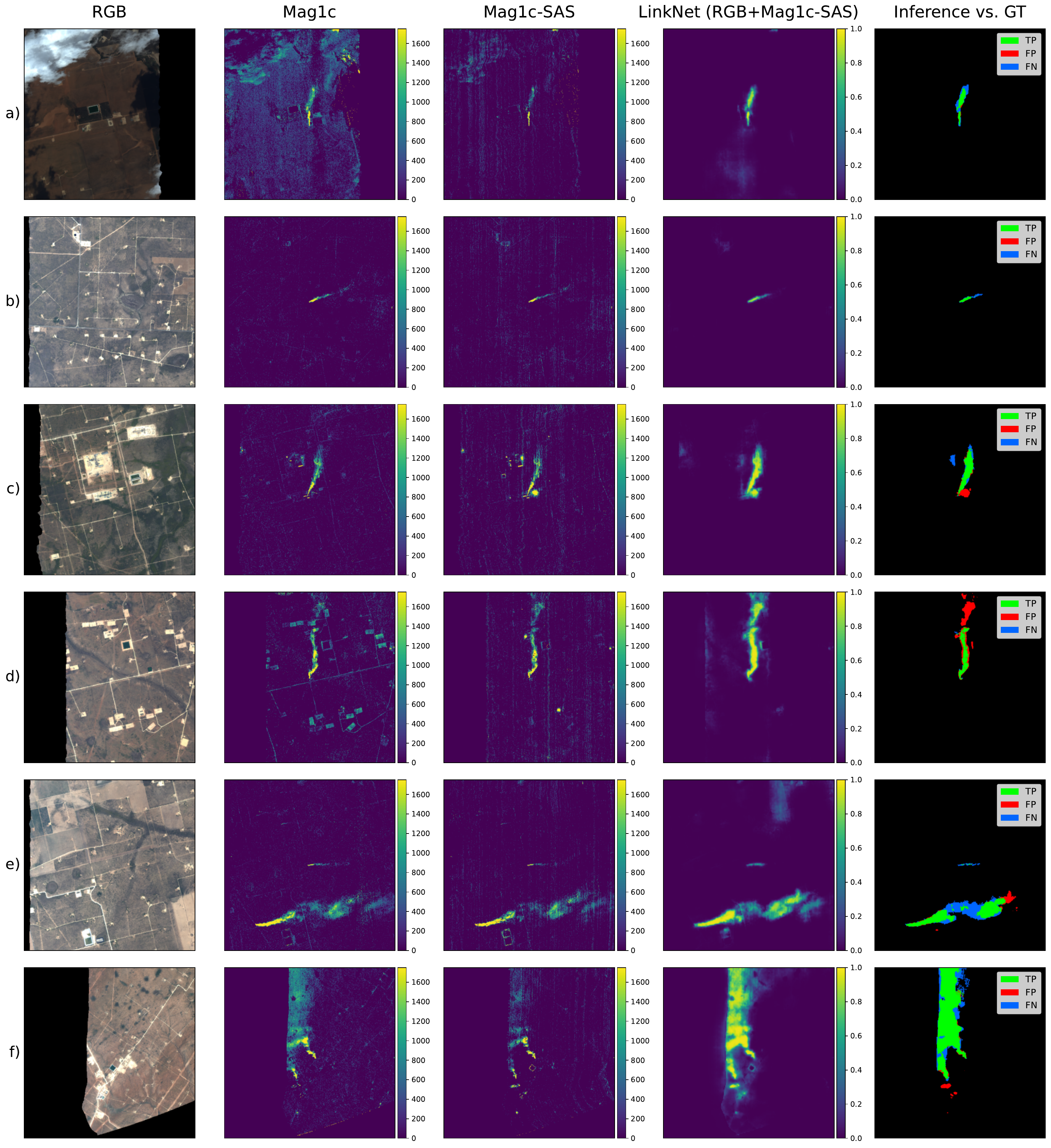}
    \caption{\textbf{Model inference.} Mag1c produces clean methane detections with smooth, cloud-like structures. Mag1c-SAS appears more fragmented and noisy, but this is effectively refined by LinkNet. In rows a) and d), Mag1c-SAS significantly reduces false positives, likely due to its larger context window. In other cases, however, it introduces noisy detections—likely false positives—potentially caused by sensor differences. Some of these are filtered out by the model, as seen in row d), while others remain, such as in row c). In row d), the potential plume tail has limited visibility in the original Mag1c and was not annotated. However, Mag1c-SAS detected it more robustly, resulting in false positives that may reflect true detections.}
    \label{fig:linknet_results}
\end{figure*}
\begin{table*}[ht!]
\caption{Results showing runtimes and metrics for 512x512 tiles with 72 channels. The total runtime is determined by adding the inference time to the optimized runtime (or, if an optimization was not implemented, the original runtime). For entries that do not mention an ML model, the morphological baseline was applied for inference. The average score and standard deviation are shown for 5 repeated training runs. The runtime is a median from 5 runs. See Section~\ref{computational-environment} for information essential to interpreting the reported runtime.} \label{tab:main-table}
\vspace{-1mm}
\begin{tabularx}{\textwidth}{@{}l*{7}{>{\centering\arraybackslash}X}@{}}
\toprule
\textbf{Method} & \textbf{Recall} & \textbf{Precision} & \textbf{F1} & \textbf{F1 - Strong} & \textbf{\makecell{Runtime [s]\\(original)}} & \textbf{\makecell{Runtime [s]\\(optimized)}} & \textbf{\makecell{Inference [s]}}\\ \midrule
Mag1c\,(original, column-wise) & 58.42 & 30.57 & 40.14 & 67.50 & 109.61 & - & + 0.34 \\ \midrule
Mag1c\,(tile-wise) & 43.71 & 22.79 & 29.96 & 51.77 & 55.19 & - & + 0.34 \\
U-Net\,[RGB+Mag1c (tile-wise)]& \textbf{65.93 ± 4.04} & 29.51 ± 6.06 & 40.29 ± 5.49 & \textbf{65.34 ± 3.4} & 55.19 & - & + 4.75 \\ \midrule
Mag1c-SAS& 52.80 & 19.44 & 28.42 & 56.34 & 1.15 & - & + 0.34 \\
U-Net\,(RGB+Mag1c-SAS)& 56.41 ± 7.0 & 34.62 ± 7.4 & 42.54 ± 6.7 & 61.38 ± 7.7 & 1.15 & - & + 4.75 \\
ACE & 25.38 & 11.09 & 15.44 & 29.58 & 13.21 & 3.73 & + 0.34 \\
U-Net\,(RGB+ACE)& 48.23 ± 4.5 & 22.01 ± 4.5 & 29.78 ± 3.8 & 48.18 ± 3.6 & 13.21 & 3.73 & + 4.75 \\
CEM & 39.92 & 11.42 & 17.76 & 39.25 & 0.49 & 0.48 & + 0.34 \\
U-Net\,(RGB+CEM) & 55.47 ± 8.4 & 23.49 ± 6.3 & 31.90 ± 4.9 & 53.55 ± 4.9 & 0.49 & 0.48 & + 4.75 \\
MF & 39.83 & 11.47 & 17.81 & 39.11 & 2.73 & 1.15 & + 0.34 \\
U-Net\,(RGB+MF)& 58.59 ± 7.0 & 21.67 ± 5.5 & 30.81 ± 4.7 & 51.94 ± 4.7 & 2.73 & 1.15 & + 4.75 \\
\midrule
LinkNet\,(RGB+Mag1c-SAS) & 51.11 ± 7.2 & \textbf{40.43 ± 6.4} & \textbf{44.44 ± 3.9} & 60.37 ± 5.1 & 1.15 & - & + 0.43 \\
LinkNet\,(RGB+CEM) & 52.72 ± 7.0 & 22.7 ± 6.5 & 31.27 ± 7.1 & 55.56 ± 6.2 & 0.49 & 0.48 & + 0.43 \\
\midrule
\bottomrule
\end{tabularx}
\vspace{-5mm}
\end{table*}
The results can be seen in Table \ref{tab:main-table}. The original Mag1c method achieves the highest F1 score for strong plumes (67.50\,\%) but has a very long runtime (109.61\,s). However, even though the runtime is measured here for a 512x512 tile for adequate comparison, the product is precomputed in the STARCOP dataset, where it was derived by column-wise application to a whole scene and tiled afterward. Thus, the original Mag1c serves more as an aspirational upper bound rather than a strict baseline.

A more appropriate baseline is the Mag1c applied tile-wise, which reduces runtime to 55.19\,s, while also experiencing a significant drop (\tildecorrect16 points) in F1 score for strong plumes. Interestingly, it was significantly enhanced by the U-Net, which decreases the drop to \tildecorrect2 points.

Among proposed approaches, the arguably strongest one is accelerated Mag1c-SAS, which achieves a drastic \tildecorrect100× speedup (1.15\,s runtime) while only sacrificing approximately 10 points of the strong plume F1 score against the original Mag1c. When combined with the U-Net, this is slightly decreased to a \tildecorrect6 points.

Among classical target detection methods, ACE exhibits the lowest detection performance while also having the longest computational time, making it the least preferable option. In contrast, CEM has the shortest runtime (0.48\,s) but also suffers a significant drop in accuracy. CEM and MF scores are very similar, which makes sense since applying CEM to whitened data (i.e., after subtracting the band means at each pixel) is equivalent to MF. Bypassing the data whitening step also makes it the simplest, thus the fastest filter. Notably, its F1 score for strong plumes improved by \tildecorrect14 points when using the U-Net.
\begin{figure}[h]
\centerline{\includegraphics[width=0.5\textwidth]{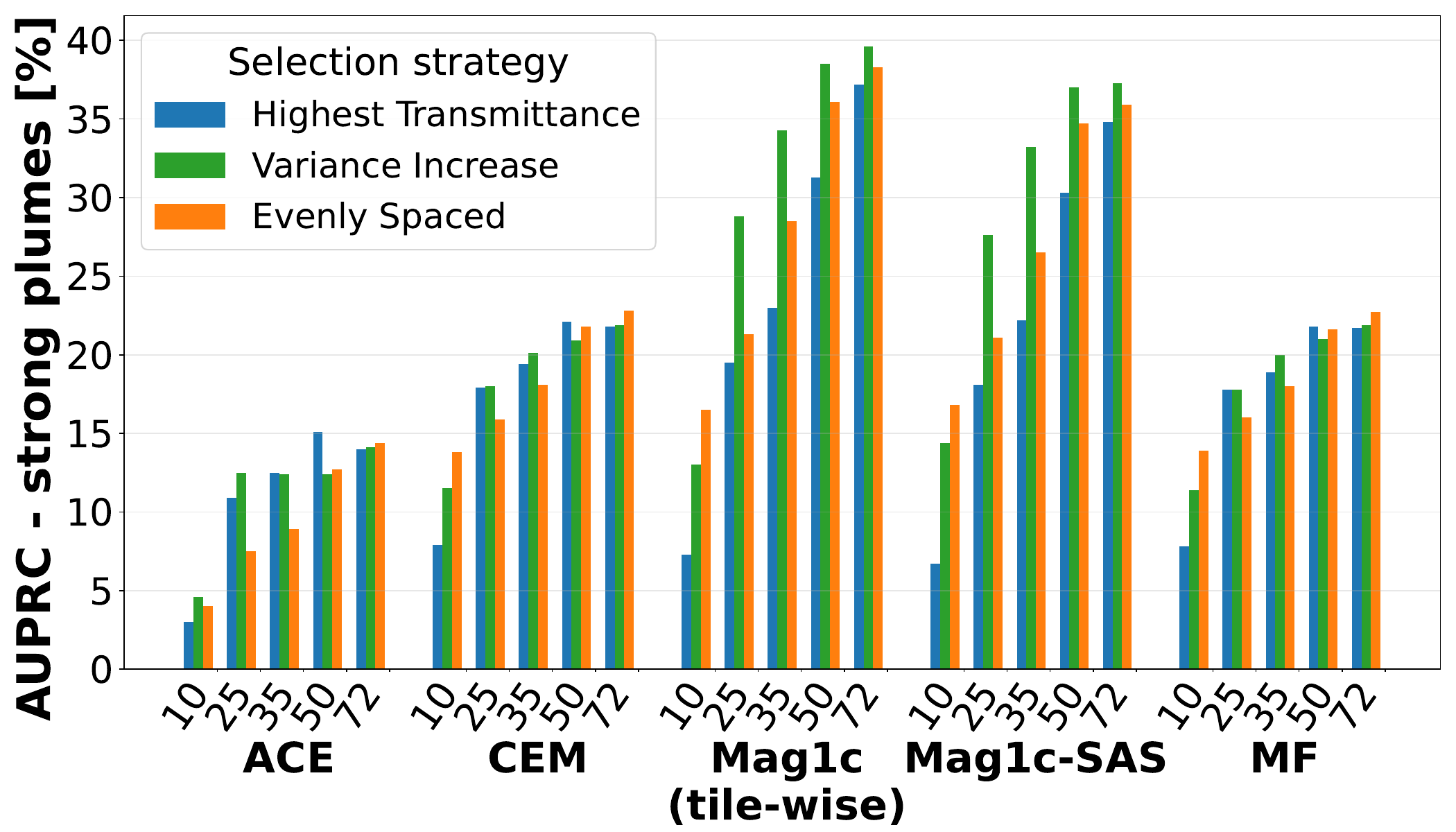}}
\caption{Band selection strategies benchmark.}
\label{select-strategy}
\vspace{-4mm}
\end{figure}

\begin{figure}[h]
\centerline{\includegraphics[width=0.5\textwidth]{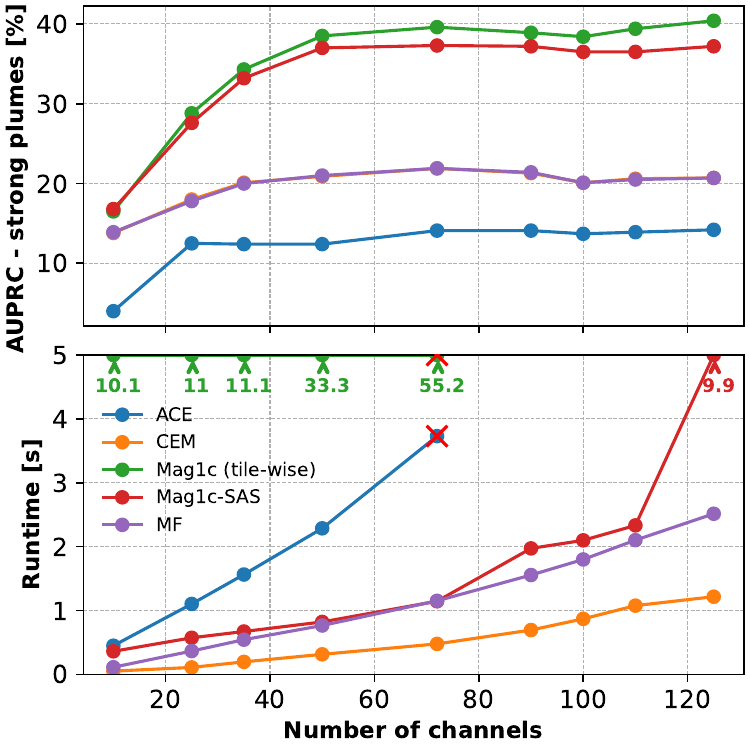}}
\caption{Runtime and AUPRC for strong plumes across different channel numbers. Red crosses indicate the last valid run, as the algorithm ran out of memory for higher channel counts.}
\label{channel-n}
\vspace{-4mm}
\end{figure}

Based on these results, we selected the two most promising candidates—CEM and Mag1c-SAS—and trained on them a more lightweight LinkNet architecture in place of the original U-Net, reducing inference time from 4.75\,s to 0.43\,s. LinkNet\,(RGB+CEM) achieved a slight improvement of around 2 points in F1 score for strong plumes, with the total runtime decreasing from 5.24\,s to 0.92\,s. LinkNet\,(RGB+Mag1c-SAS) experienced only a minor drop of approximately 1 point in F1 score for strong plumes, while the total runtime was reduced from 5.9\,s to 1.58\,s. Interestingly, this constellation received the overall highest F1 score for \textit{all} plumes. The model inference can be seen in Figure \ref{fig:linknet_results}.

\section{Results - Band selection}

All the results in Table \ref{tab:main-table} were computed using 72 spectral channels. To further explore potential runtime reductions, we investigated three band selection strategies, as illustrated in Figure \ref{select-strategy}. For every filter, except ACE, the results indicate that for a very small number of channels (N=10), the optimal strategy is to select the N channels evenly spaced in the methane-sensitive region (\tildecorrect2100-2500\,nm). For a larger number of channels, the best approaches differ; for classical filters, it bounces between the Highest transmittance and Variance Increase, often with negligible difference. But for Mag1c filters, the obvious winner is the Variance Increase strategy.

Also, it seems that using more than 50 channels brings only a negligible improvement, so we examined this for even higher channel counts to verify whether the difference between 50 and 72 wasn’t just a local effect. For this experiment, the Variance Increase selection strategy was used for all channel counts except for N=10, where the evenly spaced strategy was employed. The results are in Figure \ref{channel-n}, which suggests that the optimal number of bands truly lies between 50 and 72. Until 50 channels, the score improvements are great; improvement from 50 to 72 channels is only marginal, while beyond 72 channels, the increase is practically non-existent. Meanwhile, both runtime and memory consumption continue to rise, indicating diminishing returns in performance relative to computational cost.

\section{Discussion}

Our research involved benchmarking various filters and developing accelerated versions for methane detection. Among these, Mag1c-SAS proved promising, delivering fast and quite precise results for strong plumes when paired with a lightweight LinkNet model. The simpler CEM detector showed even greater potential for speed optimization, though with a slightly larger trade-off in accuracy.

We also extensively tested multiple band selection methods. For classic and generally less accurate methods, the impact of these strategies was minimal. However, with the more precise Mag1c algorithms, band selection made a clear difference: selecting bands that maximized the variance in transmittance was most effective. This is likely because with greater variance, the overall methane transmittance function can probably be better approximated. Our findings also suggest that the ideal number of channels for STARCOP data lies around 50, as fewer channels significantly reduced methane detection capability, and more added computational complexity.

These fast algorithms show relatively strong performance, with only a small drop in methane detection compared to the much slower Mag1c. This difference may even be slightly underestimated, as the STARCOP labels used for evaluation were created by humans based on Mag1c outputs, which are themselves imperfect. Nonetheless, the absolute performance remains modest, with F1 scores for strong plumes peaking at around 60\,\% when combined with LinkNet. So, as an alternative to our filter-based approach is to forego methane enhancement product computation entirely and instead use end-to-end machine learning (ML) models, as exemplified by \cite{Ruzicka2025HyperspectralViTs}. While such ML methods can potentially extract more information from raw hyperspectral data, leading to higher accuracy from the same sensor, our approach offers potentially greater generalizability across different sensors. Furthermore, our accelerated algorithms could be adapted for detecting other spectral signatures, extending this work beyond methane detection.

This means our algorithms could be deployed on satellites without extensive pre-launch effort or training, ready for immediate use. They could serve as an initial, broad detection system before being replaced within a few months by more accurate, end-to-end ML systems, which could be fine-tuned on the collected orbital data. To facilitate this transition and further boost acceleration, future research could focus on rewriting these algorithms into a common, onboard-compatible format. For example, since large libraries like PyTorch are impractical for spacecraft, converting the presented algorithms to frameworks like DaCe~\cite{dace}, which exports NumPy-like code directly for CPU and FPGA, presents a viable alternative.

For an end-to-end approach, our discoveries regarding the band selection could also be utilized to decrease the number of input channels while keeping the accuracy high. However, this needs to be tested as it's plausible that these strategies will perform differently for deep learning end-to-end approaches. DL inherent unpredictability and variation between training runs could mean that they might not rely on the same correlation-like principles favoring high variance as our filter methods. For example, ML models could potentially learn alternative "indices-like" approaches, such as subtracting channels with high absolute transmittance from neighboring low absolute transmittance channels.

Overall, this research demonstrates fast and almost out-of-the-box usable algorithms, potentially capable of detecting any spectral target on board low-power satellites. Combined with band selection insights, it offers important steps toward a highly generalizable, fast, and computationally efficient hyperspectral segmentation system, currently aimed at methane, but with potential for broader applications.

\section*{Acknowledgments}
The work of J.H. and R.P. on this article was co-financed under project no. FW09020069 with state support from the Technology Agency of the Czech Republic (TA CR) within the TREND programme.

\section*{Preprint notice}
This is a preprint of a paper accepted for the \href{https://atpi.eventsair.com/edhpc-2025/}{EDHPC 2025 Conference}.

\bibliographystyle{IEEEtran}
\bibliography{bib}

\end{document}